\documentclass[10pt,twocolumn,letterpaper]{article}

\usepackage{cvpr}
\usepackage{times}
\usepackage{epsfig}
\usepackage{graphicx}
\usepackage{amsmath}
\usepackage{amssymb}
\usepackage[noend]{algpseudocode}
\usepackage{algorithm}
\usepackage{enumitem}
\usepackage{gensymb}

\usepackage[pagebackref=true,breaklinks=true,letterpaper=true,colorlinks,bookmarks=false]{hyperref}

\cvprfinalcopy 

\def\cvprPaperID{408} 

\ifcvprfinal\fi
\begin{document}

\title{Cross-Domain Self-supervised Multi-task Feature Learning\\using Synthetic Imagery}

\author{Zhongzheng Ren and Yong Jae Lee\\
University of California, Davis\\
{\tt\small \{zzren, yongjaelee\}@ucdavis.edu}
}

\maketitle

\begin{abstract}
In human learning, it is common to use multiple sources of information \emph{jointly}.  However, most existing feature learning approaches learn from only a single task.  In this paper, we propose a novel multi-task deep network to learn generalizable high-level visual representations. Since multi-task learning requires annotations for multiple properties of the same training instance, we look to synthetic images to train our network.  To overcome the domain difference between real and synthetic data, we employ an unsupervised feature space domain adaptation method based on adversarial learning.  Given an input synthetic RGB image, our network simultaneously predicts its surface normal, depth, and instance contour, while also minimizing the feature space domain differences between real and synthetic data.  Through extensive experiments, we demonstrate that our network learns more transferable representations compared to single-task baselines.  Our learned representation produces state-of-the-art transfer learning results on PASCAL VOC 2007 classification and 2012 detection.
\end{abstract}

\vspace{-0.15in}
\section{Introduction}

In recent years, deep learning has brought tremendous success across various visual recognition tasks~\cite{fcn, fastrcnn, xie-iccv15}.  A key reason for this phenomenon is that deep networks trained on ImageNet~\cite{imagenet} learn \emph{transferable representations} that are useful for other related tasks.  However, building large-scale, annotated datasets like ImageNet~\cite{imagenet} is extremely costly both in time and money.  Furthermore, while benchmark datasets (e.g., MNIST~\cite{mnist}, Caltech-101~\cite{caltech101}, Pascal VOC~\cite{pascal}, ImageNet~\cite{imagenet}, MS COCO~\cite{mscoco}) enable breakthrough progress, it is only a matter of time before models begin to overfit and the next bigger and more complex dataset needs to be constructed.  The field of computer vision is in need of a more scalable solution for learning general-purpose visual representations.

Self-supervised learning is a promising direction, of which there are currently three main types.  The first uses visual cues within an image as supervision such as recovering the input from itself~\cite{denoise,autoencoder}, color from grayscale~\cite{colorization,split_brain}, equivariance of local patchs~\cite{count}, or predicting the relative position of spatially-neighboring patches~\cite{jigsaw,carl-iccv15}.  The second uses external sensory information such as motor signals~\cite{pulkit,dinesh-iccv2015} or sound~\cite{sound,look} to learn image transformations or categories. The third uses motion cues from videos~\cite{xiaolong,dinesh-cvpr2016,ishan-eccv16,deepak-2017}.  Although existing methods have demonstrated exciting results, these approaches often require delicate and cleverly-designed tasks in order to force the model to learn semantic features.  Moreover, most existing methods learn only a \emph{single task}.  While the model could learn to perform really well at that task, it may in the process lose its focus on the actual intended task, i.e., to learn high-level semantic features.  Recent self-supervised methods that do learn from multiple tasks either require a complex model to account for the potentially large differences in input data type (e.g., grayscale vs. color) and tasks (e.g., relative position vs. motion prediction)~\cite{carl-iccv17} or is designed specifically for tabletop robotic tasks and thus has difficulty generalizing to more complex real-world imagery~\cite{lerrel}.



\begin{figure}[t!]
\centering
\includegraphics[width=\columnwidth]{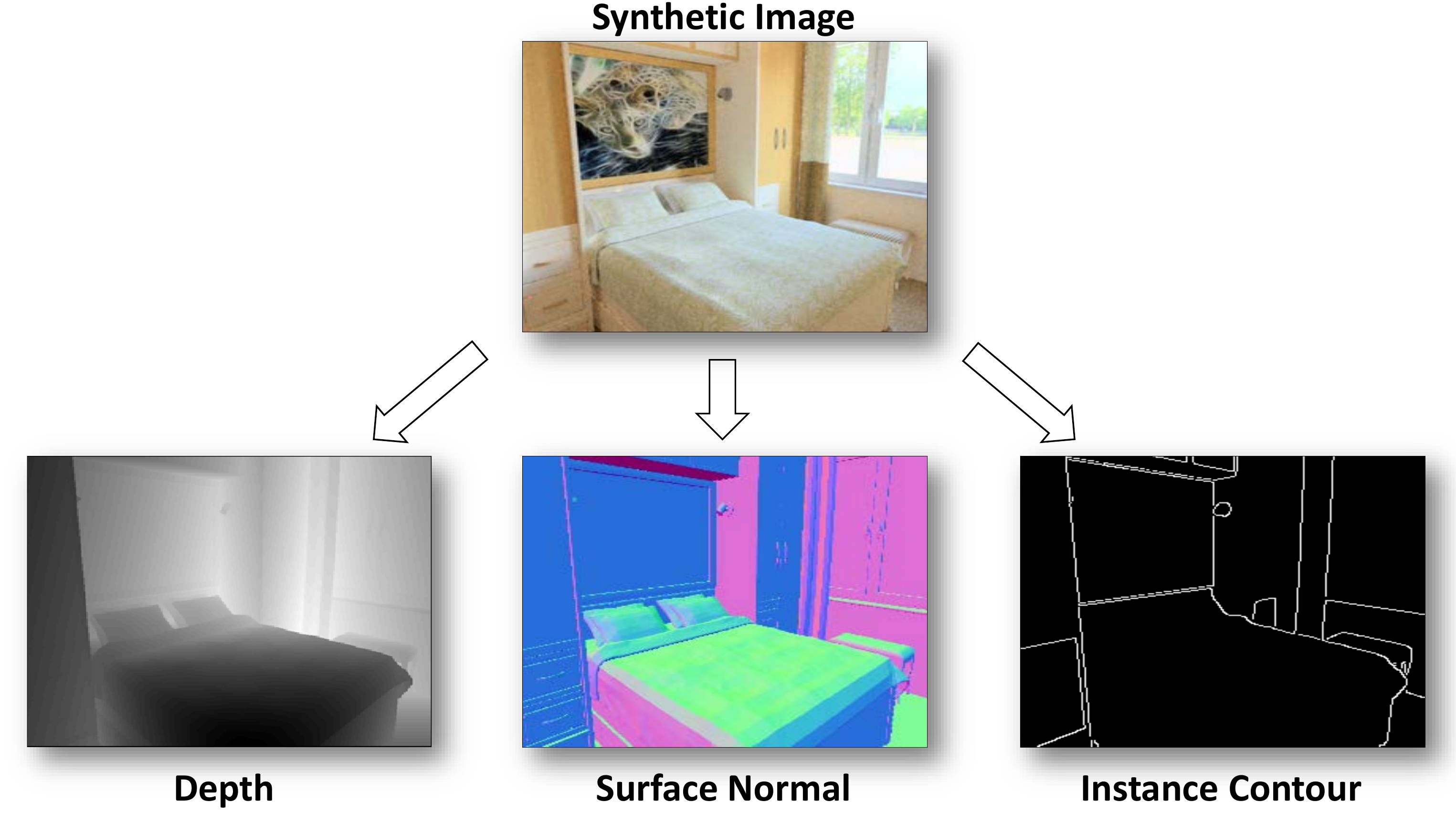}
\caption{\textbf{Main idea.} A graphics engine can be used to easily render realistic synthetic images together with their various physical property maps.  Using these images, we train a self-supervised visual representation learning algorithm in a multi-task setting that also adapts its features to real-world images.}
\label{figure:concept}
\vspace*{-0.15in}
\end{figure}

In human learning, it is common to use multiple sources of information \emph{jointly}.  Babies explore a new object by looking, touching, and even tasting it; humans learn a new language by listening, speaking, and writing in it.  We aim to use a similar strategy for visual representation learning.  Specifically, by training a model to jointly learn several complementary tasks, we can force it to learn general features that are not overfit to a single task and are instead useful for a variety of tasks.  However, multi-task learning using natural images would require access to different types of annotations (e.g., depth~\cite{eigen-iccv2015}, surface normal~\cite{eigen-iccv2015,cross-stitch}, segmentations~\cite{cross-stitch}) for each image, which would be both expensive and time-consuming to collect.

Our main idea is to instead use \emph{synthetic images} and their various \emph{free} annotations for visual representation  learning.  Why synthetic data?  First, computer graphics (CG) imagery is more realistic than ever and is only getting better over time.  Second, rendering synthetic data at scale is easier and cheaper compared to collecting and annotating photos from the real-world.  Third, a user has full control of a virtual world, including its objects, scenes, lighting, physics, etc.  For example, the global illumination or weather condition of a scene can be changed trivially.  This property would be very useful for learning a robust, invariant visual representation since the \emph{same scene} can be altered in various ways \emph{without changing the semantics}.  Finally, the CG industry is huge and continuously growing, and its created content can often be useful for computer vision researchers. For example,~\cite{gta-intel} demonstrated how the GTA-V~\cite{gta5} game can be used to quickly generate semantic segmentation labels for training a supervised segmentation model.

Although synthetic data provides many advantages, it can be still challenging to learn general-purpose features applicable to real images.  First, while synthetic images have become realistic, it's still not hard to differentiate them from real-world photos; i.e., there is a domain difference that must be overcome.  To tackle this, we propose an unsupervised feature-level domain adaptation technique using adversarial training, which leads to better performance when the learned features are transferred to real-world tasks. Second, any semantic category label must still be provided by a human annotator, which would defeat the purpose of using synthetic data for self-supervised learning.  Thus, we instead leverage other \emph{free} physical cues to learn the visual representations.  Specifically, we train a network that takes an image as input and predicts its depth, surface normal, and instance contour maps. We empirically show that learning to predict these mid-level cues forces the network to also learn transferable high-level semantics.


\vspace{-10pt}
\paragraph{Contributions.}  Our main contribution is a novel self-supervised multi-task feature learning network that learns from synthetic imagery while adapting its representation to real images via adversarial learning.  We demonstrate through extensive experiments on ImageNet and PASCAL VOC that our multi-task approach produces visual representations that are better than alternative single-task baselines, and highly competitive with the state-of-the-art.
\section{Related work}
\vspace{-3pt}
\paragraph{Synthetic data for vision.}
CAD models have been used for various vision tasks such as 2D-3D alignment~\cite{bansal-cvpr2016,Aubry14}, object detection~\cite{Peng_2015_ICCV}, joint pose estimation and image-shape alignment~\cite{haosu_sig14,Huang_sig}. Popular datasets include the Princeton Shape Benchmark~\cite{princeten_shape}, ShapeNet~\cite{shapenet}, and SUNCG~\cite{suncg}. Synthetic data has also begun to show promising usage for vision tasks including learning optical flow~\cite{flow-game}, semantic segmentation~\cite{gta-intel,synthia, Shafaei}, video analysis~\cite{virtual-kitti}, stereo~\cite{unrealstero}, navigation~\cite{yuke-icra}, and intuitive physics~\cite{uetorch,galileo,newtonian}.  In contrast to these approaches, our work uses synthetic data to learn general-purpose visual representations in a self-supervised way.  

\vspace{-10pt}
\paragraph{Representation learning.}  Representation learning has been a fundamental problem for years; see Bengio \etal ~\cite{bengio_rl} for a great survey. Classical methods such as the autoencoder~\cite{autoencoder, denoise} learn compressed features while trying to recover the input image. Recent self-supervised approaches have shown promising results, and include recovering color from a grayscale image (and vice versa)~\cite{colorization,split_brain,gustav-2017}, image inpainting~\cite{context-encoder}, predicting the relative spatial location or equivariance relation of image patches~\cite{jigsaw,carl-iccv15,count}, using motion cues in video~\cite{xiaolong,dinesh-cvpr2016,ishan-eccv16,deepak-2017}, and using GANs~\cite{adversarial_feature}.  Other works leverage non-visual sensory data to predict egomotion between image pairs~\cite{pulkit,dinesh-iccv2015} and sound from video~\cite{sound,look}.  In contrast to the above works, we explore the advantage of using \emph{multiple} tasks.

While a similar multi-task learning idea has been studied in~\cite{carl-iccv17,lerrel,xiaolong-iccv17}, each have their drawbacks.  In~\cite{carl-iccv17}, four very different tasks are combined into one learning framework.  However, because the tasks are very different in the required input data type and learning objectives, each task is learned one after the other rather than simultaneously and special care must be made to handle the different data types. In~\cite{lerrel}, a self-supervised robot learns to perform different tasks and in the process acquires useful visual features. However, it has limited transferability because the learning is specific to the tabletop robotic setting. Finally,~\cite{xiaolong-iccv17} combines the tasks of spatial location prediction~\cite{carl-iccv15} and motion coherence~\cite{xiaolong}, by first initializing with the weights learned on spatial location prediction and then continuing to learn via motion coherence (along with transitive relations acquired in the process).  Compared to these methods, our model is relatively simple yet generalizes well, and learns all tasks simultaneously.


\vspace{-10pt}
\paragraph{Domain adaptation.} To overcome dataset bias, \emph{visual} domain adaptation was first introduced in~\cite{saenko-2010}.  Recent methods using deep networks align features by minimizing some distance function across the domains~\cite{tzeng_iccv15, ganin2015}.  GAN~\cite{gan} based pixel-level domain adaptation methods have also gained a lot of attention and include those that require paired data~\cite{pix2pix} as well as unpaired data~\cite{CycleGAN,DiscoGAN,CoGAN}.

Domain adaptation techniques have also been used to adapt models trained on synthetic data to real-world tasks~\cite{apple, dilip-cvpr17}.  Our model also minimizes the domain gap between real and synthetic images, but we perform domain adaptation in feature space similar to~\cite{tzeng-cvpr17, DANN}, whereby a domain discriminator learns to distinguish the domains while the learned representation (through a generator) tries to fool the discriminator.  To our knowledge, our model is the first to adapt the features learned on synthetic data to real images for self-supervised feature learning.  


\vspace{-10pt}
\paragraph{Multi-task learning.} Multi-task learning~\cite{mtl} has been used for a variety vision problems including surface normal and depth prediction~\cite{eigen-iccv2015,eigen_nips14}, semantic segmentation~\cite{cross-stitch}, pose estimation~\cite{poseactionrcnn}, robot manipulation~\cite{lerrel_icra17,lerrel}, and face detection~\cite{zhanpeng}.  Kokkinos~\cite{ubernet} introduces a method to jointly learn low-, mid-, and high-level vision tasks in a unified architecture.  Inspired by these works, we use multi-task learning for self-supervised feature learning. We demonstrate that our multi-task learning approach learns better representations compared to single-task learning.

\section{Approach}
\label{sec:approach}

We introduce our self-supervised deep network which jointly learns multiple tasks for visual representation learning, and the domain adaptor which minimizes the feature space domain gap between real and synthetic images. Our final learned features will be transferred to real-world tasks.

\subsection{Multi-task feature learning}

\begin{figure}[t!]
\centering
\includegraphics[scale = 0.57]{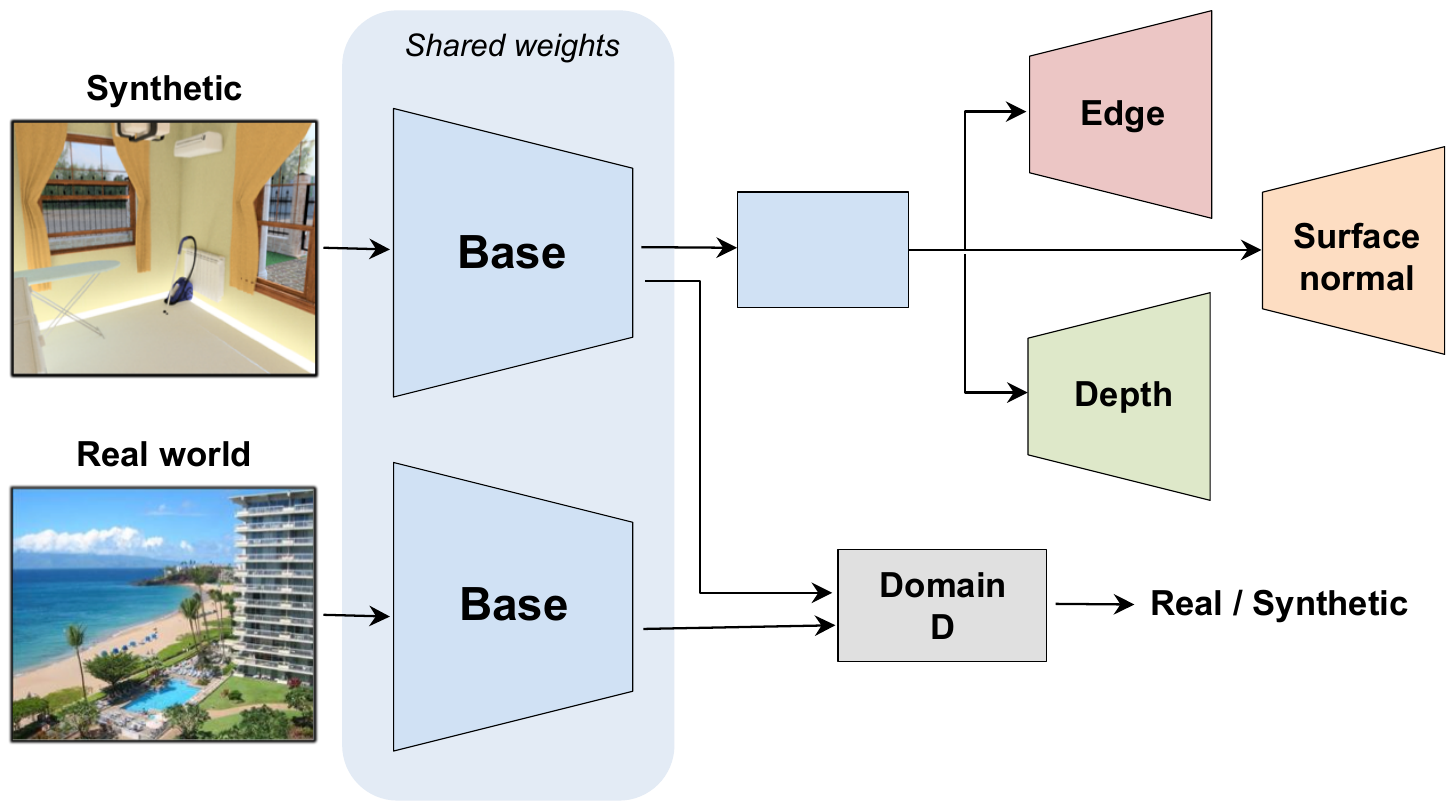}
\caption{\textbf{Network architecture.} The upper net takes a synthetic image and predicts its depth, surface normal, and instance contour map. The bottom net extracts features from a real-world image.  The domain discriminator D tries to differentiate real and synthetic features. The learned blue modules are used for transfer learning on real-world tasks.}
\label{figure:archi}
\end{figure}

To learn general-purpose features that are useful for a variety of tasks, we train our network to simultaneously solve three different tasks.  Specifically, our network takes as input a single synthetic image and computes its corresponding instance contour map, depth map, and surface normal map, as shown in Fig.~\ref{figure:archi}.  

\vspace{-10pt}
\paragraph{Instance contour detection.}  We can easily extract instance-level segmentation masks from synthetic imagery.  The masks are generated from pre-built 3D models, and are clean and accurate.  However, the tags associated with an instance are typically noisy or inconsistent (e.g., two identical chairs from different synthetic scenes could be named `chair1' and `furniture2').  Fixing these errors (e.g., for semantic segmentation) would require a human annotator, which would defeat the purpose of self-supervised learning.

We therefore instead opt to extract edges from the instance-level segmentation masks, which alleviates the issues with noisy instance labels.  For this, we simply run the canny edge detector on the segmentation masks.  Since the edges are extracted from instance-level segmentations, they correspond to \emph{semantic} edges (i.e., contours of objects) as opposed to low-level edges.  Fig.~\ref{figure:concept} shows an example; notice how the edges within an object, texture, and shadows are ignored.  Using these semantic contour maps, we can train a model to ignore the low-level edges within an object and focus instead on the high-level edges that separate one object from another, which is exactly what we want in a high-level feature learning algorithm.

More specifically, we formulate the task as a binary semantic edge/non-edge prediction task, and use the class-balanced sigmoid cross entropy loss proposed in~\cite{xie-iccv15}:
\begin{equation*}
\vspace{-3pt}
\resizebox{\hsize}{!}{$L_e(E) = -\beta \sum_{i} \log P(y_i=1 | \theta)\; - (1-\beta)\sum_{j} \log P(y_j=0 | \theta)$}
\label{eq:edge}
\vspace{-2pt}
\end{equation*}
where $E$ is our predicted edge map, $E'$ is the ground-truth edge map, $\beta = |E'_-|/|E'_- + E'_+|$, and $|E'_-|$ and $|E'_+|$ denote the number of ground-truth edges and non-edges, respectively, $i$ indexes the ground-truth edge pixels, $j$ indexes the ground-truth background pixels, $\theta$ denotes the network parameters, and $P(y_i=1|\theta)$ and $P(y_j=0|\theta)$ are the predicted probabilities for a pixel corresponding to an edge and background, respectively.

\vspace{-10pt}
\paragraph{Depth prediction.} Existing feature learning methods mainly focus on designing `pre-text' tasks such as predicting the relative position of spatial patches~\cite{carl-iccv15,jigsaw} or image in-painting~\cite{context-encoder}.  The underlying physical properties of a scene like its depth or surface normal have been largely unexplored for learning representations.  The only exception is the work of~\cite{pixelnet}, which learns using surface normals corresponding to real-world images. (In Sec.~\ref{sec:results}, we demonstrate that our multi-task approach using synthetic data leads to better transferable representations.)

Predicting the depth for each pixel in an image requires understanding high-level semantics about objects and their relative placements in a scene; it requires the model to figure out the objects that are closer/farther from the camera, and their shape and pose.  While real-world depth imagery computed using a depth camera (e.g., the Kinect) can often be noisy, the depth map extracted from a synthetic scene is clean and accurate.  To train the network to predict depth, we follow the approach of~\cite{eigen_nips14}, which compares the predicted and ground-truth log depth maps of an image $Q = \log~Y$ and $Q' = \log~Y'$, where $Y$ and $Y'$ are the predicted and ground-truth depth maps, respectively.  Their scale-invariant depth prediction loss is:
\begin{equation*}
\resizebox{.65 \columnwidth}{!} {\resizebox{\hsize}{!}
{ $L_d(Q) = \frac{1}{n}\sum_i d_i^{2} - \frac{1}{2n^{2}} \sum_{i,j} d_i d_j$}}
\label{eq:ldd}
\end{equation*}
where $i$ indexes the pixels in an image, $n$ is the total number of pixels, and $d = Q - Q'$ is the element-wise difference between the predicted and ground-truth log depth maps.  The first term is the L2 difference and the second term tries to enforce errors to be consistent with one another in their sign.

\vspace{-10pt}
\paragraph{Surface normal estimation.}
Surface normal is highly related to depth, and previous work~\cite{eigen-iccv2015,eigen_nips14} show that combing the two tasks can help both.  We use the inverse of the dot product between the ground-truth and the prediction as the loss~\cite{eigen-iccv2015}:
\begin{equation*}
\resizebox{.44 \columnwidth}{!}{\resizebox{\hsize}{!}
{ $L_{s}(S)	= - \frac{1}{n}\sum_i  S_i \cdot S_i' $}}
\label{eq:l1l2}
\end{equation*}
where $i$ indexes the pixels in an image, $n$ is the total number of pixels, $S$ is the predicted surface normal map, and $S'$ is the ground-truth surface normal map.

\subsection{Unsupervised feature space domain adaptation}

While the features learned above on multiple tasks will be more general-purpose than those learned on a single task, they will not be directly useful for real-world tasks due to the domain gap between synthetic and real images.  Thus, we next describe how to adapt the features learned on synthetic images to real images.

Since our goal is to learn features in a self-supervised way, we cannot assume that we have access to any task labels for real images.  We therefore formulate the problem as \emph{unsupervised} domain adaptation, where the goal is to minimize the domain gap between synthetic $x_i\in X$ and real $y_j\in Y$ images.  We follow a generative adversarial learning (GAN)~\cite{gan} approach, which pits a generator and a discriminator against each other. In our case, the two networks learn from each other to minimize the domain difference between synthetic and real-world images so that the features learned on synthetic images can generalize to real-world images, similar to~\cite{DANN,apple, dilip-cvpr17,tzeng-cvpr17}.  Since the domain gap between our synthetic data and real images can be potentially huge (especially in terms of high-level semantics), we opt to perform the adaptation at the feature-level~\cite{DANN,tzeng-cvpr17} rather than at the pixel-level~\cite{apple, dilip-cvpr17}.

Specifically, we update the discriminator and generator networks by alternating the following two stages.  In the first stage, given a batch of synthetic images $\textbf{x} = \{x_i\}$ and a batch of real images $\textbf{y} = \{y_j\}$, the generator $B$ (base network in Fig.~\ref{figure:archi}) computes features $z_{x_i} = B(x_i)$ and $z_{y_j} = B(y_j)$ for each synthetic image ${x_i}$ and real image ${y_j}$, respectively.  The domain discriminator $D$ then updates its parameters $\phi_D$ by minimizing the following binary cross-entropy loss:
\begin{equation*}
\resizebox{.97 \columnwidth}{!}{\resizebox{\hsize}{!}
{$L_D(\phi_D|z_{\textbf{x}},z_{\textbf{y}}) = -\sum\nolimits_i \log(D(z_{x_i})) - \sum\nolimits_j \log(1-D(z_{y_j}))$}}
\vspace{-2pt}
\end{equation*}
where we assign $1, 0$ labels to synthetic and real images $x_i, y_j$, respectively.

In the second stage, we fix $D$ and update the generator $B$ as well as the tasks heads $H$ for the three tasks.  Specifically, the parameters $\phi_B, \phi_H$ are updated jointly using:
\begin{multline*}
L_{BH}(\phi_B,\phi_H|z_{\textbf{x}}) = - \sum\nolimits_i \log(1-D(z_{x_i})) \\
				 + \lambda_e L_{e}(E_{x_i}) + \lambda_d L_{d}(Q_{x_i}) + \lambda_s L_{s}(S_{x_i}),
\end{multline*}
where $L_{e}(E_{x_i}),L_{d}(Q_{x_i}), L_{s}(S_{x_i})$ are the losses for instance contour, depth, and surface normal prediction for synthetic image $x_i$, respectively, and $\lambda_e, \lambda_d, \lambda_s$ are weights to scale their gradients to have similar magnitude. $L_{BH}$ updates $B$ so that $D$ is fooled into thinking that the features extracted from a synthetic image are from a real image, while also updating $H$ so that the features are good for instance contour, depth, and surface normal prediction.

\begin{algorithm}[t]
\footnotesize
\caption{Multi-task Adversarial Domain Adaptation} \label{trainalg}
\begin{algorithmic}[1]
      \Require{Synthetic images $X$, real images $Y$, max iteration $T$}
      \Ensure{Domain adapted base network $B$}
        \For{$t = 1$ to ${T}$}
            	\State Sample a batch of synthetic images $\textbf{x} = \{x_i\}$
      		\State Sample a batch of real images $\textbf{y} = \{y_j\}$
      		\State Extract feature for each image: $z_{x_i} = B(x_i), z_{y_j} = B(y_j)$
      		\State Keep $D$ frozen, update $B,H$ through $L_{BH}(\phi_B,\phi_H|z_{\textbf{x}})$
      		\State Keep $B$ frozen, update $D$ through $L_D(\phi_D|z_{\textbf{x}},z_{\textbf{y}})$
        \EndFor
\end{algorithmic}
\end{algorithm}

Our training process is summarized in Alg.~\ref{trainalg}. Note that we do not directly update the generator $B$ using any real images; instead the real images only directly update $D$, which in turn forces $B$ to produce more domain-agnostic features for synthetic images.  We also tried updating $B$ with real images (by adding  $-\sum_j \log(D(z_{y_j}))$ to $L_{BH}$), but this did not result in any improvement.  Once training converges, we transfer $B$ and finetune it on real-world tasks like ImageNet classification and PASCAL VOC detection.

\begin{figure*}[t!]
\centering
\includegraphics[width=\textwidth]{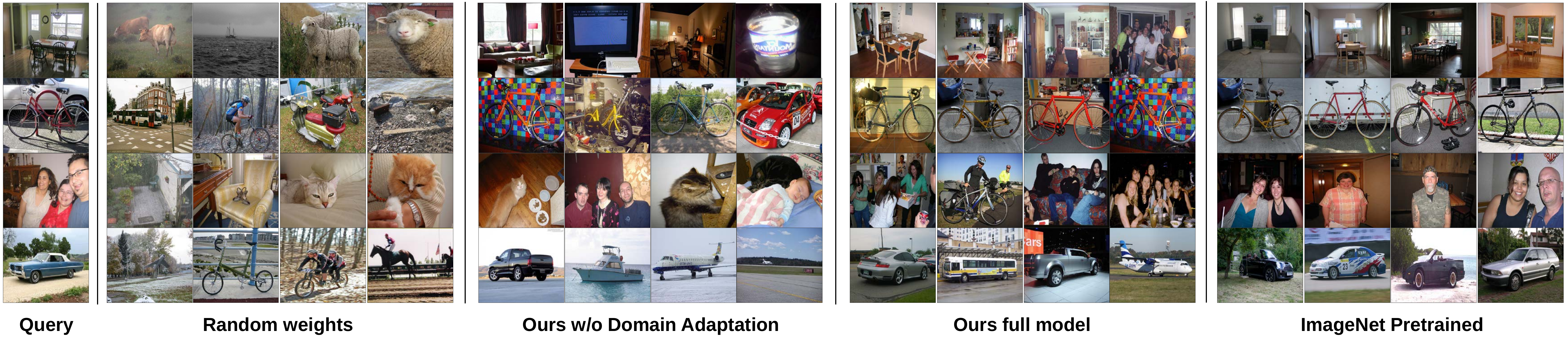}
\caption{Nearest neighbor retrieval results. The first column contains the query images. We show the four nearest neighbors of a randomly initialized AlexNet, our model without domain adaptation, our model with domain adaptation, and ImageNet pre-trained AlexNet.}
\label{figure:nn}
\vspace*{-0.1in}
\end{figure*}

\subsection{Network architecture}
\label{sec:architecture}

Our network architecture is shown in Fig.~\ref{figure:archi}.  The blue base network consists of convolutional layers, followed by ReLU nonlinearity and BatchNorm~\cite{bn}. The ensuing bottleneck layers (middle blue block) consist of dilated convolution layers~\cite{fisher} to enlarge the receptive field.  In our experiments, the number of layers and filters in the base and bottleneck blocks follow the standard AlexNet~\cite{alex-net} model to ensure a fair comparison with existing self-supervised feature learning methods (e.g.,~\cite{carl-iccv15,colorization,split_brain,count}).  The task heads (red, green, and orange blocks) consist of deconvolution layers, followed by ReLU and BatchNorm~\cite{bn}. Finally, the domain discriminator is a $13\times13$ patch discriminator~\cite{pix2pix}, which takes `conv5' features from the base network.  Exact architecture details are provided in the Appendix.

Empirically, we find that minimizing the domain shift in a mid-level feature space like `conv5' rather than at a lower or higher feature space produces the best transfer learning results.  In Sec.~\ref{sec:results}, we validate the effect of adaptation across different layers.

\section{Results}
\label{sec:results}

In this section, we evaluate the quality and transferability of the features that our model learns from synthetic data.  We first produce qualitative visualizations of our learned conv1 filters, nearest neighbors obtained using our learned features, and learned task predictions on synthetic data.  We then evaluate on transfer learning benchmarks: fine-tuning the features on PASCAL VOC classification and detection, and freezing the features learned from synthetic data and then training a classifier on top of them for ImageNet classification.  We then conduct ablation studies to analyze the different components of our algorithm.  Finally, we evaluate our features on NYUD surface normal prediction.

\subsection{Experimental setup}

\paragraph{Architecture}
As described in Sec.~\ref{sec:architecture}, we set our base network to use the same convolutional and pooling layers as AlexNet~\cite{alex-net} (the blue blocks in Fig.~\ref{figure:archi}) to ensure a fair comparison with existing self-supervised approaches~\cite{colorization,adversarial_feature,carl-iccv15,xiaolong,dinesh-cvpr2016,pulkit,sound,xiaolong-iccv17}.  We set our input to be grayscale by randomly duplicating one of the RGB channels three times since it can lead to more robust features~\cite{count,carl-iccv15,xiaolong}.

\vspace{-10pt}
\paragraph{Dataset}
We use Places365~\cite{places} as the source of real images for domain adaptation, which contains 1.8 million images. For synthetic images, we combine SUNCG~\cite{suncg} and SceneNet RGB-D~\cite{scenenet} to train our network.  Both datasets come with depth maps for each synthetic image, and we compute instance contour maps from the provided instance masks.  For surface normal, we use the ground-truth maps provided by~\cite{xiaolong} for SceneNet~\cite{scenenet} and those provided by SUNCG~\cite{suncg}.

\begin{figure}[t!]
\centering
\includegraphics[scale = 0.26]{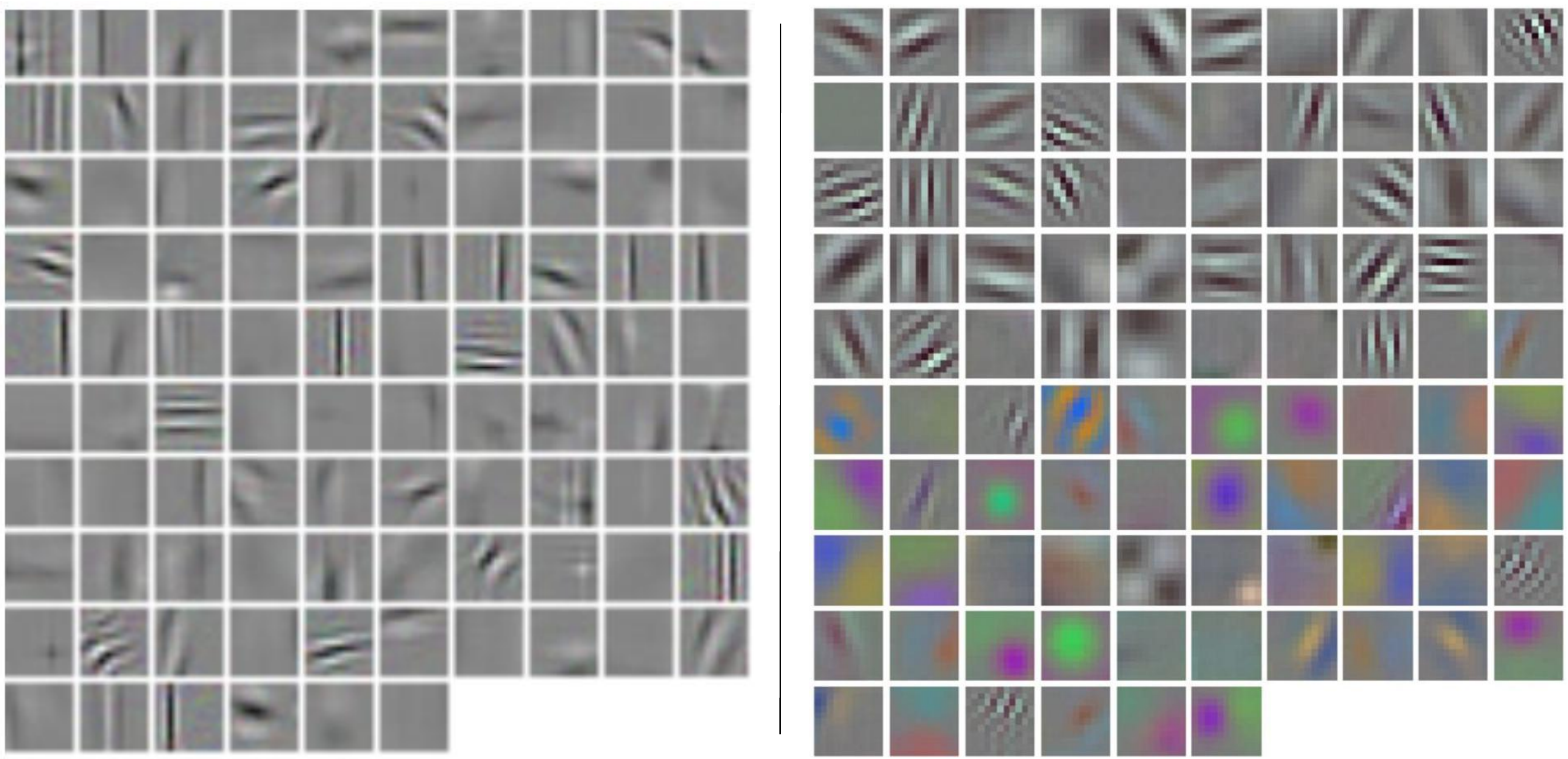}
\caption{(\textbf{left}) The conv1 filters learned using our model on SUNCG and SceneNet. (\textbf{right}) The conv1 filters learned on ImageNet.  While not as sharp as those learned on ImageNet, our model learns gabor-like conv1 filters.}
\label{figure:conv1}
\vspace*{-0.1in}
\end{figure}

\begin{figure*}[t!]
\centering
\includegraphics[width=0.95\textwidth]{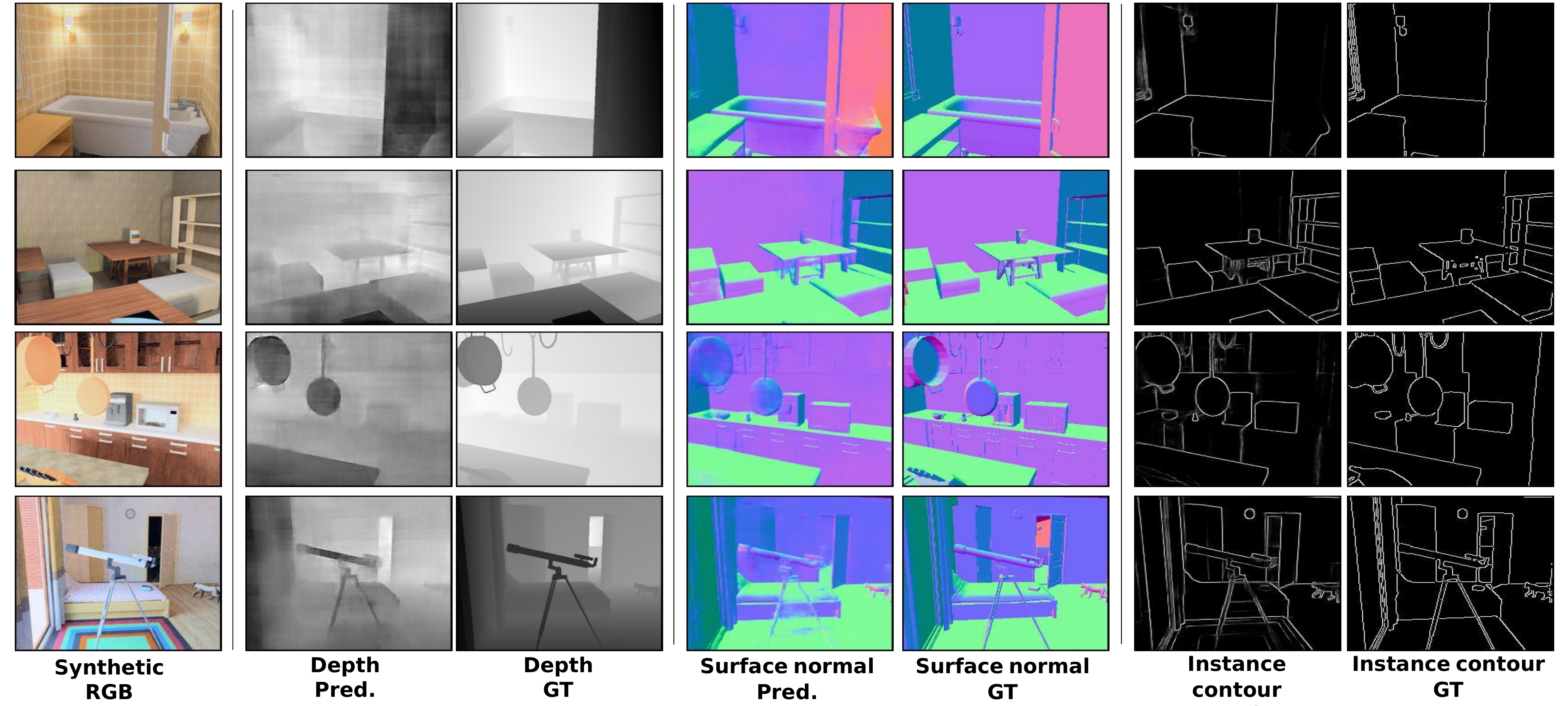}
\caption{Representative examples of our model's depth, surface normal, and instance contour predictions on unseen SUNCG~\cite{suncg} images.  Our network produces predictions that are sharp and detailed, and close to the ground-truth.}
\label{figure:quali}
\vspace*{-0.1in}
\end{figure*}

\subsection{Qualitative analysis without finetuning}

\paragraph{Nearest neighbor retrieval}

We first perform nearest neighbor retrieval experiments on the PASCAL VOC 2012 trainval dataset.  For this experiment, we compare a randomly initalized AlexNet, ImagenNet pretrained AlexNet, our model without domain adaptation, and our full model with domain adaptation.  For each model, we extract conv5 features for each VOC image and retrieve the nearest neighbors for each query image.

Fig.~\ref{figure:nn} shows example results.  We make several observations: (1) Both our full model and model without domain adaptation produces better features than randomly initialized features. (2) Since many of the ImageNet objects are not present in our synthetic dataset, our model is unable to distinguish between very similar categories but instead retrieves them together (e.g., cars, buses, and airplanes as the neighbor of query car). (3) Our full model performs better than our model without domain adaptation when there are humans or animals in the query images. This is likely because although these categories are never seen in our synthetic training set, they are common in Places~\cite{places} which we use for adaptation. (4) Compared to a pre-trained ImageNet~\cite{imagenet} model, our full model is less discriminative and prefers to capture images with more objects in the image (e.g., third row with humans). This may again be due to Places~\cite{places} since it is a scene dataset rather than an object-centric dataset like ImageNet.  Overall, this result can be seen as initial evidence that our pre-trained model can capture high-level semantics on real-world data.

\vspace{-10pt}
\paragraph{Conv1 filter visualization}
In Fig.~\ref{figure:conv1}, we visualize the conv1 features learned on synthetic data.  While not as sharp as those learned on ImageNet~\cite{imagenet}, our model learns conv1 features that resemble gabor-like filters.  Since we always convert our input image to gray scale, our network does not learn any color blob filters.

\vspace{-10pt}
\paragraph{Learned task prediction visualization}
We next show how well our model performs on the tasks that it is trained on.  Fig.~\ref{figure:quali} shows our model's depth, surface normal, and instance contour predictions on unseen SUNCG~\cite{suncg} images.  Overall, our predictions are sharp and clean, and look quite close to the ground-truth.  Note that these are representative predictions and we only sampled these because they contain interesting failure cases.  For example, in the first row there is a transparent glass door. Our network failures to capture the semantic meaning of a glass door and instead tries to predict the bathtub's surface normal and contours behind it.  In the third row, our network fails to correctly predict the pan and pot's depth and surface normals due to ambiguity in 3D shape. This indicates that our network can struggle when predicting very detailed 3D properties. Similar results can been seen in the fourth row with the telescope body and legs.  Finally, in the last row, there is a door whose inside is too dark to see. Therefore, our network predicts it as a wall but the ground-truth indicates there is actually something inside it.


These visualizations illustrate how well our network performs on each `pre-text' task for feature learning.  The better our model performs on these tasks, the better transferable features it is likely to get.  In the remainder of the experiments, we demonstrate that this is indeed the case, and also provide quantitative evaluations on the surface normal `pre-text' task in Sec.~\ref{sec:nyud-surfacenormal} where we fine-tune our network for surface normal estimation on NYUD~\cite{nyud}.

\subsection{Transfer learning}

\paragraph{Pascal VOC classification and detection}

We first evaluate on VOC classification following the protocol in~\cite{data_init}.  We transfer the learned weights from our network (blue blocks Fig.~\ref{figure:archi}) to a standard AlexNet~\cite{alex-net} and then re-scale the weights using~\cite{data_init}.  We then fine-tune our model's weights on VOC 2007 trainval and test on VOC 2007 test.  Table~\ref{pascal} second column, shows the results.  Our model outperforms all previous methods despite never having directly used any real images for pre-training (recall that the real images are only used for domain adaptation).  In contrast, the existing methods are all trained on real images or videos.  While previous research has mainly shown that synthetic data can be a good supplement to real-world imagery~\cite{gta-intel, synthia}, this result indicates the promise of directly using synthetic data and its free annotations for self-supervised representation learning.

We next test VOC detection accuracy using the Fast-RCNN~\cite{fastrcnn} detector.  We test two models: (1) finetuning on VOC 2007 trainval and testing on VOC 2007 test data; (2) finetuning on VOC 2012 train and testing on VOC 2012 val data.  Table~\ref{pascal}, right two columns show the results.  Our models obtain the second best result on VOC 2007 and the best result on 2012.  These results on detection verify that our learned features are robust and are able to generalize across different high-level tasks.  More importantly, it again shows that despite only using synthetic data, we can still learn transferable visual semantics.

\begin{table}[t!]
\centering
 \footnotesize{
\begin{tabular}{c| c | c c }
\hline
Dataset								& 07 	& 07 	& 12 	\\
Tasks 								& CLS. 	& DET. 	& DET.	\\
\hline
ImageNet \cite{alex-net} 	 	& 79.9 	& 56.8 	& 56.5 	\\
\hline
Gaussian 							& 53.4 	& 41.3 	& - 		\\
Autoencoder	\cite{data_init}			& 53.8	& 41.9	& -		\\	
Krahenbuel \etal	\cite{data_init}  	& 56.6 	& 45.6 	& 42.8 	\\
\hline
Ego-equivariance \cite{dinesh-iccv2015} 	& -		& 41.7 	& -	 	\\
Egomotion \cite{pulkit} 			& 54.2	& 43.9 	& -	 	\\
context-encoder \cite{context-encoder} 	& 56.5 & 44.5 & \\
BiGAN \cite{adversarial_feature} 	& 58.6	& 46.2	& 44.9	\\
sound \cite{sound} 				& 61.3	& -	 	& 42.9	\\
flow \cite{deepak-2017} 			& 61 	& 52.2	& 48.6	\\
motion \cite{xiaolong}			& 63.1	& 47.2	& 43.5	\\
clustering \cite{clustering}		& 65.3	& 49.4	& -		\\
context \cite{data_init} 			& 65.3	& 51.1 	& 49.9 	\\
colorization \cite{colorization} 	& 65.9	& 46.9	& 44.5	\\
jigsaw \cite{jigsaw}				& 67.6 	& \textbf{53.2}  	& -  \\
splitbrain \cite{split_brain} 	& 67.1	& 46.7	& 43.8	\\
counting \cite{count} 			& 67.7	& 51.4	& -	\\
\hline
Ours 						& \textbf{68.0}	& 52.6	& \textbf{50.0}		\\
\hline
\end{tabular}
}
\caption{Transfer learning results on PASCAL VOC 2007 classification and VOC 2007 and 2012 detection. We report the best numbers for each method reported in~\cite{data_init, split_brain, count}.}
\label{pascal}
\vspace*{-0.1in}
\end{table}

\vspace{-10pt}
\paragraph{ImageNet classification}
We next evaluate our learned features on ImageNet classification~\cite{imagenet}. We freeze our network's pre-trained weights and train a multinomial logistic regression classifier on top of each layer from conv1 to conv5 using the ImageNet classification training data.  Following~\cite{split_brain}, we bilinearly interpolate the feature maps of each layer so that the resulting flattened features across layers produce roughly equal number of dimensions.

Table~\ref{table:imgnet} shows the results. Our model shows improvement over the different data initialization methods (Gaussian and Kr\"ahenb\"uhl~\etal~\cite{data_init}), but underperforms compared to the state-of-the-art. This is understandable since existing self-supervised approaches~\cite{carl-iccv15,adversarial_feature,context-encoder,colorization} are trained on ImageNet, which here is also the test dataset.  Our model is instead trained on synthetic indoor images, which can have quite different high-level semantics and thus has never seen most of the ImageNet categories during training (e.g., there are no dogs in SUNCG). Still, it outperforms~\cite{context-encoder} and performs similarly to~\cite{colorization} up through conv4, which shows that the learned semantics on synthetic data can still be useful for real-world image classification.

\subsection{Ablation studies}

We next perform ablation studies to dissect the contribution of the different components of our model.  For this, we again use the PASCAL VOC classification and detection tasks for transfer learning.


\begin{table}[t!]
\centering
\resizebox{\columnwidth}{!}{
\begin{tabular}{c| c c c c c}
\hline
method 									& conv1 	& conv2 	& conv3 	& conv4 	& conv5 \\
\hline
ImageNet~\cite{alex-net} 					& 19.3 	& 36.3 	& 44.2 	& 48.3 	& 50.5	\\
\hline
Gaussian 								& 11.6 	& 17.1 	& 16.9 	& 16.3 	& 14.1 	\\
Kr\"ahenb\"uhl~\etal \cite{data_init} 		& 17.5 	& 23.0 	& 24.5 	& 23.2 	& 20.6 	\\
\hline
context \cite{carl-iccv15}					& 16.2 	& 23.3	& 30.2  	& 31.7 	& 29.6 	\\
BiGAN \cite{adversarial_feature}			& 17.7 	& 24.5	& 31.0  	& 29.9 	& 28.0 	\\
context-encoder \cite{context-encoder}		& 14.1 	& 20.7	& 21.0  	& 19.8 	& 15.5 	\\
colorization \cite{colorization}			& 12.5  & 24.5 	& 30.4 	& 31.5 	& 30.3 	\\
jigsaw \cite{jigsaw}						& \textbf{18.2} 	& 28.8  	& 34.0  	& 33.9  	& 27.1 	\\
splitbrain \cite{split_brain}		& 17.7	& 29.3  	& \textbf{35.4}  & \textbf{35.2} 	& \textbf{32.8} 	\\
counting \cite{count}				& 18.0	& \textbf{30.6}  	& 34.3  	& 32.5  	& 25.7 	\\
\hline
Ours	& 16.5	& 27.0	& 30.5	& 30.1  	& 26.5	\\
\hline
\end{tabular}
}
\caption{Transfer learning results on ImageNet~\cite{imagenet}. We freeze the weights of our model and train a linear classifier for ImageNet classification~\cite{imagenet}.  Our model is trained purely on synthetic data while all other methods are trained on ImageNet~\cite{imagenet} (without labels).  Despite the domain gap, our model still learns useful features for image classification.}
\label{table:imgnet}
\vspace*{-0.1in}
\end{table}

\vspace{-10pt}
\paragraph{Does multi-task learning help in learning semantics?}
We first analyze whether multi-task learning produces more transferable features compared to single-task learning.  Table~\ref{table:ablation}, first four rows show the transfer learning results of our final multi-task model (`3 tasks') versus each single-task model (`Edge', `Depth', `Surf.').  Our multi-task model outperforms all single-task models on both VOC classification and detection, which demonstrates that the tasks are complementary and that multi-task learning is beneficial for feature learning.

\vspace{-10pt}
\paragraph{Does domain adaptation help? If so, on which layer should it be performed?}
Table~\ref{table:ablation}, rows 5-8 show the transfer learning results after applying domain adaptation in different layers (i.e., in Fig.~\ref{figure:archi}, which layer's features will go into the domain discriminator).  We see that domain adaptation helps when performed on conv5 and conv6\footnote{Since our pre-text tasks are pixel prediction tasks, we convert fc6-7 of AlexNet into equivalent conv6-7 layers.}, which verifies that there is indeed a domain difference between our synthetic and real images that needs to be addressed.  For example, on VOC classification, performing domain adaptation on conv5 results in 67.4\% accuracy vs.~65.6\% without domain adaptation.  Interestingly, we see a slight decrease in performance from conv5 to conv6  across all tasks (rows 7 \& 8).  We hypothesize that this drop in performance is due to the biases in the synthetic and real-world image datasets we use; SUNCG and SceneNet are both comprised of indoor scenes mostly with man-made objects whereas Places is much more diverse and consists of indoor and outdoor scenes with man-made, natural, and living objects.  Thus, the very high-level semantic differences will be hard to overcome, and domain adaptation becomes difficult at the very high layers.

We also see that it actually hurts to perform domain adaptation at a very low-layer like conv1.  The low performance on conv1 is likely due to the imperfect rendering quality of the synthetic data that we use.  Many of the rendered images from SUNCG~\cite{suncg} are a bit noisy.  Hence, if we take the first layer's conv1 features for domain adaptation, it is easy for the discriminator to overfit to this artifact, which causes low-level differences from real images.  Indeed, we find that the conv1 filters learned in this setting are quite noisy, and this leads to lower transfer learning performance.  By performing the domain-adaptation at a higher-level, we find that the competition between the discriminator and generator better levels-out, leading to improved transfer learning performance.  Overall, performing domain adaptation in between the very low and very high layers, such as conv5, results in the best performance.

\begin{table}[t]
\centering
 \footnotesize{
\begin{tabular}{c c c |c c c}
\hline
Task 	& Adaptation	 & \#data & 07-C & 07-D &12-D\\
\hline
Edge 	& -		& 0.5M	& 63.9  	& 46.9	& 44.8	\\
Depth 	& -		& 0.5M	& 61.9  	& 48.9	& 45.8	\\
Surf. 	& -		& 0.5M	& 65.3  	& 48.2	& 45.4	\\
3 tasks	& -		& 0.5M	& \textbf{65.6}  	&  \textbf{51.3}	&  \textbf{47.2}	\\
\hline
3 tasks	& conv1		& 0.5M	& 61.9	& 48.7	 & 46   \\
3 tasks	& conv4		& 0.5M	& 63.4  	& 49.5	 & 46.3 \\
3 tasks	& conv5		& 0.5M	&  \textbf{67.4}  	&  \textbf{52.0}	 &  \textbf{49.2} \\
3 tasks	& conv6		& 0.5M	& 66.9  	& 51.5	 & 48.2 \\
\hline
3 tasks	& conv5	Bi-fool	& 0.5M	& 66.2  	& 51.3	& 48.5 	\\
\hline
3 tasks	& conv5	  	& 1.5M	&  \textbf{68.0}  	&  \textbf{52.6}	&  \textbf{50.0}	\\
\hline
\end{tabular}
}
\caption{Ablation study results. We evaluate the impact of multi-task learning, feature space domain adaptation, and amount of data on transfer learning.  All of these factors contribute together to make our model learn transferable visual features from large-scale synthetic data.}
\label{table:ablation}
\vspace*{-0.1in}
\end{table}


\vspace{-10pt}
\paragraph{Does more data help?}
The main benefit of self-supervised or unsupervised learning methods is their scalability since they do not need any manually-labeled data.  Thus, we next evaluate the impact that increasing data size has on feature learning.  Specifically, we increase the size of our synthetic dataset from 0.5 million images to 1.5 million images.  From Table~\ref{table:ablation}, we can clearly see that having more data helps (`3task conv5' model, rows 7 vs.~10).  Specifically, both classification and detection performance improves by 0.5-0.6\% points.

\vspace{-10pt}
\paragraph{Does fooling the discriminator both ways help?}
Since both of our real and synthetic images go through one base network, in contrast to standard GAN architectures, during the generator update we can fool the discriminator in both ways (i.e., generate synthetic features that look real and real image features that look synthetic).  As seen in Table~\ref{table:ablation}, row 9, fooling the discriminator in this way hurts the performance slightly, compared to only generating synthetic features that look real (row 7), but is still better than no domain adaptation (row 4).  One likely reason for this is that updating the generator to fool the discriminator into thinking that a real image feature is synthetic does not directly help the generator produce good features for the synthetic depth, surface normal, and instance contour tasks (which are ultimately what is needed to learn semantics).  Thus, by fooling the discriminator in both ways, the optimization process becomes unnecessarily tougher.  This issue could potentially be solved using stabilizing methods such as a history buffer~\cite{apple}, which we leave for future study.

\begin{table}[t!]
\centering
\resizebox{\columnwidth}{!}{
\begin{tabular}{c|c|c c|c c c}
\hline
\multicolumn{2}{c}{} & \multicolumn{2}{|c|}{Lower the better} & \multicolumn{3}{|c}{Higher the better} \\
\hline
GT & Methods 	& Mean	& Median		& $11.25\degree$  & $22.5\degree$  &  $30\degree$ \\
\hline
\cite{eigen-iccv2015}  & Zhang \etal~\cite{zhang2016} 		    & 22.1 & 14.8 	& \textbf{39.6} 	& 65.6 & 75.3  \\
\cite{eigen-iccv2015} & Ours & \textbf{21.9}	& \textbf{14.6}	 & 39.5	& \textbf{66.7}	& \textbf{76.5} \\
\hline
\cite{ladicky}	& Wang \etal~\cite{xiaolong-iccv17}  	& 26.0 	& 18.0 	& 33.9 	& 57.6 	& 67.5 \\
\cite{ladicky}	& Ours 	& \textbf{23.8}	& \textbf{16.2} 	& \textbf{36.6} 	& \textbf{62.0} 	& \textbf{72.9} \\
\hline
\end{tabular}
}
\caption{Surface normal estimation on the NYUD~\cite{nyud} test set.}
\label{table:normal}
\vspace*{-0.1in}
\end{table}

\subsection{Surface normal on NYUD}
\label{sec:nyud-surfacenormal}

Finally, we evaluate our model's transfer learning performance on the NYUD~\cite{nyud} dataset for surface normal estimation.  Since one of our pre-training tasks is surface normal estimation, this experiment also allows us to measure how well our model does in learning that task. We use the standard split of 795 images for training and 654 images for testing.  The evaluation metrics we use are the \textbf{Mean, Median, RMSE} error and percentage of pixels that have angle error less than $\textbf{11.25\degree, 22.5\degree, and~30\degree}$  between the model predictions and the ground-truth predictions.  We use both the ground-truths provided by~\cite{ladicky} and~\cite{eigen-iccv2015}.

We compare our model with the self-supervised model of~\cite{xiaolong-iccv17}, which pre-trains on the combined tasks of spatial location prediction~\cite{carl-iccv15} and motion coherence~\cite{xiaolong}, and the supervised model trained with synthetic data~\cite{zhang2016}, which pre-trains on ImageNet classification and SUNCG surface normal estimation.  For this experiment, we use an FCN~\cite{fcn} architecture with skip connections similar to~\cite{zhang2016} and pre-train on 0.5 million SUNCG synthetic images on joint surface normal, depth, and instance contour prediction.

Table~\ref{table:normal} shows the results.  Our model clearly outperforms~\cite{xiaolong-iccv17}, which is somewhat expected since we directly pre-train on surface normal estimation as one of the tasks, and performs slightly better than~\cite{zhang2016} on average.  Our model still needs to adapt from synthetic to real images, so our good performance likely indicates that (1) our model performs well on the pre-training tasks (surface normal estimation being one of them) and (2) our domain adaptation reduces the domain gap between synthetic and real images to ease fine-tuning.

\vspace{-5pt}
\section{Conclusion}
While synthetic data has become more realistic than ever before, prior work has not explored learning general-purpose visual representations from them.  Our novel cross-domain multi-task feature learning network takes a promising first step in this direction.

\vspace{-12pt}
\paragraph{Acknowledgements.} This work was supported in part by the National Science Foundation under Grant No. 1748387, the AWS Cloud Credits for Research Program, and GPUs donated by NVIDIA.

{\small
\bibliographystyle{ieee}
\bibliography{egbib}
}

\section{Appendix}

The details of our network architectures are provided here. We first introduce the AlexNet based network used for the experiments in Sections 4.2-4.4. We then describe the VGG16 based network used for surface normal prediction on NYUD in Section 4.5.

\subsection{AlexNet}
Our network details are give in Table~\ref{table:archi1}.  There are mainly 4 components in our network (recall Figure 2 in the main paper): base network (big blue blobs), bottleneck network (small blue block), task heads (orange, red, and green blocks), and domain discriminator (gray block).

Our base network takes a 227$\times$227$\times$3 image as input. The conv1 to conv5 layers are identical to those in AlexNet~\cite{alex-net}. We change the stride of pool5 from 2 to 1 to avoid losing too much spatial information, following~\cite{colorization}.  For the bottleneck block, we use a dilated convolutional layer~\cite{fisher} in fc6 to increase its receptive field.  The base and bottleneck network can be combined and converted into a standard AlexNet~\cite{alex-net}, which we use for the transfer learning experiments. During conversion, we absorb the batch normalization layer, convert the convolutional fc6-7 into full-connected layers, and rescale the cross-layer weights. All of the above operations are identical to~\cite{colorization}.

\begin{table}[b!]
\centering
\footnotesize
\begin{tabular}{c | c c c c c c}
\hline
Layer 	& S 		& C 		& KS		& St		& P	 	& D	\\
\hline
Input 	& 227	& 3		& -		& -		& -		& -	\\
\hline
conv1 	& 55		& 96		& 11		& 4		& 0		& 1	\\
pool1 	& 55		& 96		& 3		& 2		& 0		& 1	\\
conv2 	& 27		& 256	& 5		& 1		& 2		& 1	\\
pool2 	& 27		& 256	& 3		& 2		& 0		& 1	\\
conv3 	& 13		& 384	& 3		& 1		& 0		& 1	\\
conv4 	& 13		& 384	& 3		& 1		& 0		& 1	\\
conv5 	& 13		& 256	& 3		& 1		& 0		& 1	\\
pool5 	& 13		& 256	& 3		& 1		& 1		& 1	\\
\hline
fc6 		& 13		& 4096	& 6		& 1		& 5		& 2	\\
fc7	 	& 13		& 4096	& 1		& 1		& 0		& 1	\\
\hline
D1	& 13		& 256	& 3		& 2		& 0		& 1	\\
D2	& 6		& 512	& 1		& 1		& 0		& 1	\\
D3	& 6		& 1		& 1		& 1		& 0		& 1	\\
\hline
Deconv8 	& 27		& 64		& 3		& 2		& 0		& 1	\\
Deconv9 	& 55		& 64		& 3		& 2		& 0		& 1	\\
Deconv10 & 112	& 64		& 5		& 2		& 0		& 1	\\
Output & 227	& 1 or 3	& 3		& 2		& 0		& 1	\\
\hline
\end{tabular}
\caption{AlexNet based architecture. S: spatial size of output; C: number of channels; KS: kernel size; St: stride; P: padding; D: dilation. Note: fc6, fc7 are fully convolutional layers as in FCN~\cite{fcn}.}
\label{table:archi1}
\vspace*{-0.1in}
\end{table}

Three deconvolutional layers (Deconv8 - Deconv10) are used to recover the full-sized image outputs for each task.  The output layer is also deconvolutional and has three channels for surface normal prediction, and one for depth prediction and instance contour detection.

We use a patch discriminator as in~\cite{pix2pix} whose final output is a $6\times6$ feature map. There are three layers in our domain discriminator (D1 - D3), which takes as input the conv5 output. Leaky ReLU~\cite{prelu} with slope 0.2 and batch normalization comes after the convolutional layers to stabilize the adversarial training process.

\begin{table}[t!]
\centering
\footnotesize
\begin{tabular}{ c | c c c c c c}
\hline
Layer 		& S 		& C 		& KS		& St		& P	 	& D	\\
\hline
Input 		& 224	& 3		& -		& -		& -		& -	\\
\hline
conv1\_1 	& 224	& 64		& 3		& 1		& 1		& 1	\\
conv1\_2 	& 224	& 64		& 3		& 1		& 1		& 1	\\
pool1 		& 112	& 64		& 2		& 2		& 0		& 1	\\
conv2\_1 	& 112	& 128	& 3		& 1		& 1		& 1	\\
conv2\_2 	& 112	& 128	& 3		& 1		& 1		& 1	\\
pool2 		& 56		& 128	& 2		& 2		& 0		& 1	\\
conv3\_1 	& 56		& 256	& 3		& 1		& 1		& 1	\\
conv3\_2 	& 56		& 256	& 3		& 1		& 1		& 1	\\
conv3\_3 	& 56		& 256	& 3		& 1		& 1		& 1	\\
pool3 		& 28		& 256	& 2		& 2		& 0		& 1	\\
conv4\_1 	& 28		& 512	& 3		& 1		& 1		& 1	\\
conv4\_2 	& 28		& 512	& 3		& 1		& 1		& 1	\\
conv4\_3 	& 28		& 512	& 3		& 1		& 1		& 1	\\
pool4 		& 14		& 512	& 2		& 2		& 0		& 1	\\
conv5\_1 	& 14		& 512	& 3		& 1		& 1		& 1	\\
conv5\_2 	& 14		& 512	& 3		& 1		& 1		& 1	\\
conv5\_3 	& 14		& 512	& 3		& 1		& 1		& 1	\\
\hline
D1 			& 6 		& 1024	& 4		& 2		& 0		& 1	\\
D2 			& 6		& 1024	& 1		& 1		& 0		& 0	\\
D3 			& 6		& 1024	& 1		& 1		& 0		& 0	\\
\hline
Deconv1 		& 14		& 512	& 4		& 2		& 0		& 1	\\
Deconv2 		& 28		& 256	& 4		& 2		& 0		& 1	\\
Deconv3		& 56		& 128	& 4		& 2		& 0		& 1	\\
Deconv4		& 112	& 64		& 4		& 2		& 0		& 1	\\
output		& 224	& 1 or 3	& 4		& 2		& 0		& 1	\\
\hline
\end{tabular}
\caption{VGG16 based architecture. S: spatial size of output; C: $\#$ of channels; KS: kernel size; St: stride; P: padding; D: dilation.}
\label{table:archi2}
\vspace*{-0.1in}
\end{table}

\subsection{VGG16}
Our VGG16 based network has three basic components: base network, task heads, and domain discriminator as shown in Table~\ref{table:archi2}.  To save memory, unlike our AlexNet based architecture, we do not have a bottleneck network.

Our base network takes a 224$\times$224$\times$3 image as input. The conv1\_1 to conv5\_3 layers are identical to VGG16~\cite{vgg}.  To obtain accurate pixel-level predictions for the three tasks, we use skip connections between the base and task heads (we do not do this for our AlexNet architecture for fair comparison with prior feature learning work). We use ${(a \rightarrow b)}$ to denote a skip connection from the output of a to the input of b.  The skip connections in our network are {(conv2\_2 $\rightarrow$ Deconv4)}, {(conv3\_3 $\rightarrow$ Deconv3)}, and {(conv4\_3 $\rightarrow$ Deconv2)}.  Similar to our AlexNet architecture, we use a patch discriminator, leaky ReLU, and batch normalization in the three layers of the discriminator, which takes as input  conv5\_3 output features.

\end{document}